\documentclass[runningheads]{llncs}
\usepackage[T1]{fontenc}

\usepackage{graphicx}
\usepackage{amsmath}
\usepackage{wrapfig}
\usepackage{amsfonts}
\usepackage{subcaption} 
\usepackage{hyphenat}

\begin{document}
\title{Multimodal Video Emotion Recognition with Reliable Reasoning Priors}
%
%
\newcommand{\correspondingauthor}{\textsuperscript{*}}

\author{Zhepeng Wang\inst{1} \and
Yingjian Zhu \inst{2,3} \and
Guanghao Dong \inst{4} \and
Hongzhu Yi \inst{5} \and \\
Feng Chen \inst{1} \and
Xinming Wang\inst{3,2}\correspondingauthor \and
Jun Xie\inst{1} \correspondingauthor 
}

\institute{
Lenovo Research
\email{\{wangzpb,xiejun,chenfeng13\}@lenovo.com} \and
School of Artificial Intelligence, UCAS \and
Institute of Automation, CAS 
\email{\{zhuyingjian2024, wangxinming2024\}@ia.ac.cn} \and
Macau University of Science and Technology
\email{2240007837@student.must.edu.mo}\and
School of Computer Science and Technology, UCAS
}

\maketitle              
%

\renewcommand{\thefootnote}{$^{*}$} 
\setcounter{footnote}{0}                         
\footnotetext[1]{Corresponding Author.}

\begin{abstract}
This study investigates the integration of trustworthy prior reasoning knowledge from MLLMs into multimodal emotion recognition. We employ Gemini to generate fine-grained, modality-separable reasoning traces, which are injected as priors during the fusion stage to enrich cross-modal interactions. To mitigate the pronounced class-imbalance in multimodal emotion recognition, we introduce Balanced Dual-Contrastive Learning, a loss formulation that jointly balances inter-class and intra-class distributions. Applied to the MER2024 benchmark, our prior-enhanced framework yields substantial performance gains, demonstrating that the reliability of MLLM-derived reasoning can be synergistically combined with the domain adaptability of lightweight fusion networks for robust, scalable emotion recognition.
\end{abstract}

\section{Introduction}
Recent advances in reinforcement learning~\cite{guo2025deepseek} with verified reward (RLVR)~\cite{lambert2024tulu3} have accelerated the adoption of reinforcement-learning-based post-training and chain-of-thought~\cite{wei2022chain} procedures that follow a Reflection–Answer paradigm across multimodal large language models (MLLMs), gaining great traction. These substantially expand the generalization frontier beyond what standard in-context learning or supervised fine-tuning (SFT)~\cite{ouyang2022training} can achieve.

Empirical studies consistently show that embedding a CoT paradigm within MLLMs markedly improves generalization. Nevertheless, this test-time scaling~\cite{chen2024expanding} strategy also introduces considerable computational overhead during both training and inference, exceeding that of conventional multimodal fusion pipelines. Moreover, identifying the appropriate level of reasoning granularity for different task scenarios remains an open question. Focusing on multimodal video emotion recognition, this work advances two central inquiries: (i) For a well-trained reasoning model, can its strong, generalized reasoning priors be integrated into an LLM-agnostic, multimodal recognition framework? (ii) In multimodal video analysis, how can we conduct reliable and coherent reasoning over the emotional cues conveyed by both the human subjects and their surrounding context?

To address the aforementioned challenges, we propose a modified multimodal fusion framework. This approach elicits reliable reasoning priors from MLLMs and then fuses these priors into a task-specific multimodal architecture to guide subsequent training. Conceptually, this can be viewed as targeted distillation, where the MLLMs acts as a teacher model at the instructional level and the traditional lightweight networks serve as a student model at the implementation level. Such distillation elevates the performance ceiling of the student model.

Our contributions are as follows:
\begin{itemize}
    \item We distill high-level reasoning priors from MLLMs into a lightweight multimodal recognition model, enhancing generalization with minimal computational cost.
    \item We introduce a balanced contrastive strategy to address label imbalance and improve emotion class separability in multimodal feature space.
\end{itemize}

\section{Related Work}
\subsubsection{Video Emotion Recognition}
Development of multimodal fusion and LLMs brings new prospects for video emotion recognition. transformer audio-visual fusion method like AVT-CA~\cite{dhanith2025avtca} and TACFN~\cite{liu2023tacfn} which synchronises and filters cross-modal cues for new highs on CMU-MOSEI~\cite{bagherzadeh2018cmu_mosei}, RAVDESS~\cite{livingstone2018ravdess}. Vision-language prompting method like EmoVCLIP~\cite{qi2024emovclip} and SoVTP~\cite{wang2024sovtp} utilise prompt tuning and modality dropout in the multimodal fusion process. Also, Video-Emotion-OVR~\cite{lian2025ovmer} and AffectGPT~\cite{lian2025affectgpt} shift from closed-set labels to free-form textual emotion descriptions, enriching emotion expression. 
\subsubsection{Visual Reasoning Model}
As the CoT progressively propagates the notion of “System-2” reasoning~\cite{li2025system}, the research community has begun to focus more intently on inference mechanisms for large multimodal models. Visual-RFT~\cite{liu2025visual} has empirically validated the efficacy of RLVR in purely visual settings, while VCTP leverages visual chain-of-thought prompts to enhance zero-shot visual reasoning.  ~\cite{gupta2022visprog,chen2024vctp,yariv2024multipleimage} have further sought to improve the interpretability and trustworthiness of model outputs~\cite{lu2024vqadiffusion}. Nevertheless, whether such reasoning traces can be distilled into reliable prior knowledge that meaningfully guides conventional models remains an open and intriguing research question.

\section{Method}
\subsection{Emotion Recognition with Reliable Reasoning\label{sec3.1}}

Although MLLMs exhibit impressive zero-shot capabilities and strong generalisability on general-purpose benchmarks, they often struggle to achieve high discriminative accuracy on domain-specific video-emotion corpora—underscoring a sizeable domain gap between broad and specialised tasks. 

Reliable identification and categorization of human emotions have long been a focal point of sociological research. Traditionally, scholars have modelled affect by mapping facial expressions onto either the Arousal–Valence~\cite{kuppens2013relation,libkuman2004arousal} continuum or the Facial Action Coding System (FACS)~\cite{ekman1978facial}. FACS localises 44 anatomically defined Action Units (AUs), and canonical combinations of these AUs constitute six basic-emotion prototypes, providing a robust quantitative framework for emotion classification.
\begin{figure}
    \centering
    \includegraphics[width=0.8\linewidth]{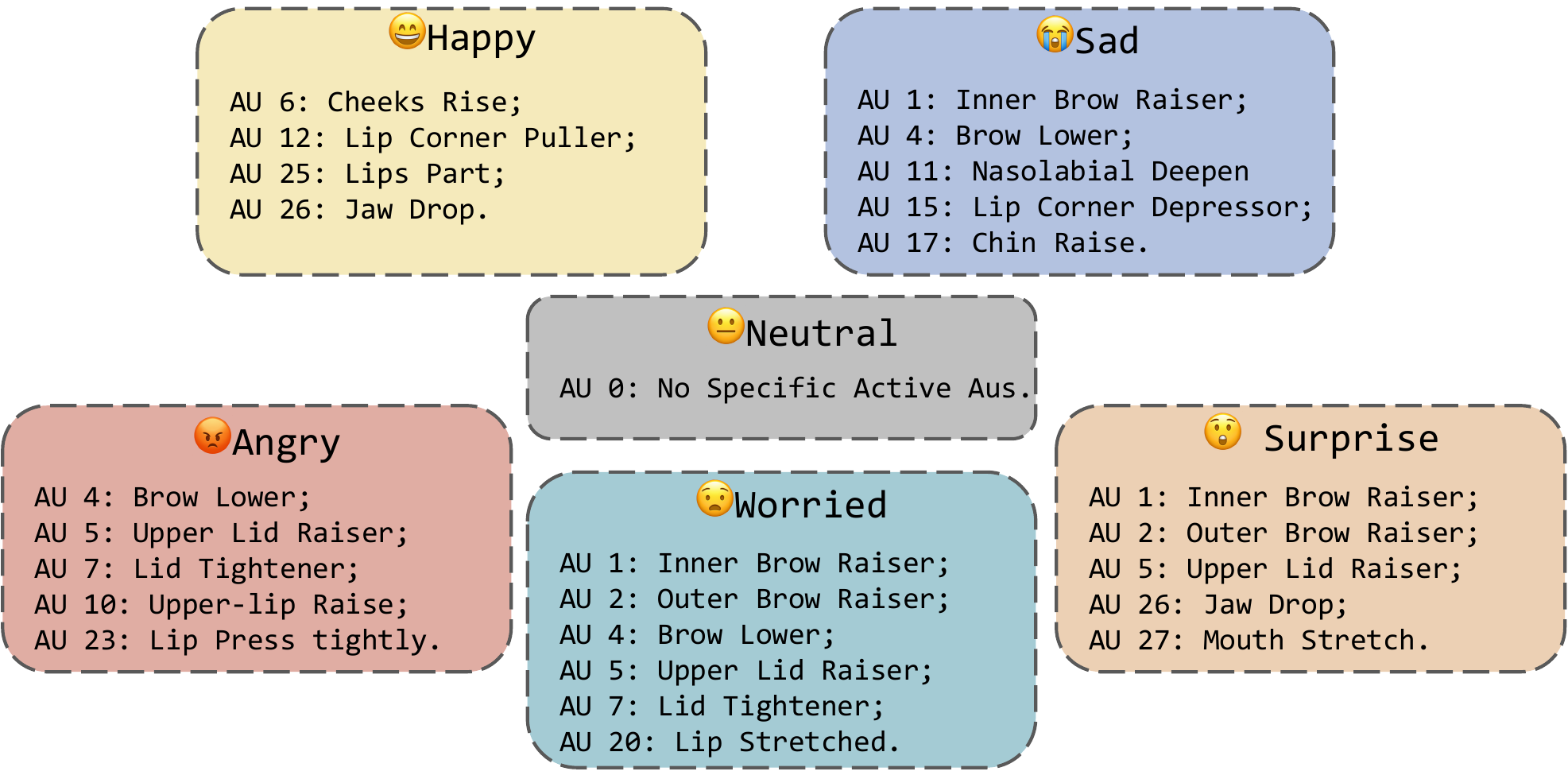}
    \caption{Mapping between Action Units and emotions}
    \label{fig:actions}
\end{figure}

Let $\mathcal{C}=\{c_1,c_2,\dots ,c_{|\mathcal{C}|}\}$ denote the set of discrete emotion classes and let $\mathcal{A}=\{\text{AU}_1,\text{AU}_2,\dots ,\text{AU}_{44}\}$
denote the complete repertoire of facial Action Units.
For every emotion label $c\in\mathcal{C}$ we have its AU-support set as
\begin{equation}
    \mathcal{S}_c \;:=\; \bigl\{\,\text{AU}_j \in \mathcal{A}\;\bigm|\;\text{AU}_j\;\text{is associated with emotion }c \bigr\}\;
\end{equation}

When performing emotion recognition, describing facial expressions through AU patterns as $I_{V}:=\{\text{AU}_i \in \mathcal{A}\}$ offers empirical support for labelling decisions, where $I$ generated from MLLMs with Instructions. 

As Gemini model family demonstrates strong multimodal reasoning and generalisation, we employ the Gemini-2.0-exp~\cite{team2023gemini} model to analyse representative video samples along three complementary dimensions: 
\begin{itemize}
    \item AU combinations extracted from facial frames $I_V$
    \item Prosodic cues derived from the audio track $I_A$;
    \item Semantic content from subtitle transcripts $I_T$.
\end{itemize}

The model further quantifies the relative contribution of video, audio, and textual channels, enabling an integrated affective judgement.
\begin{figure}
    \centering
    \includegraphics[width=0.9\linewidth]{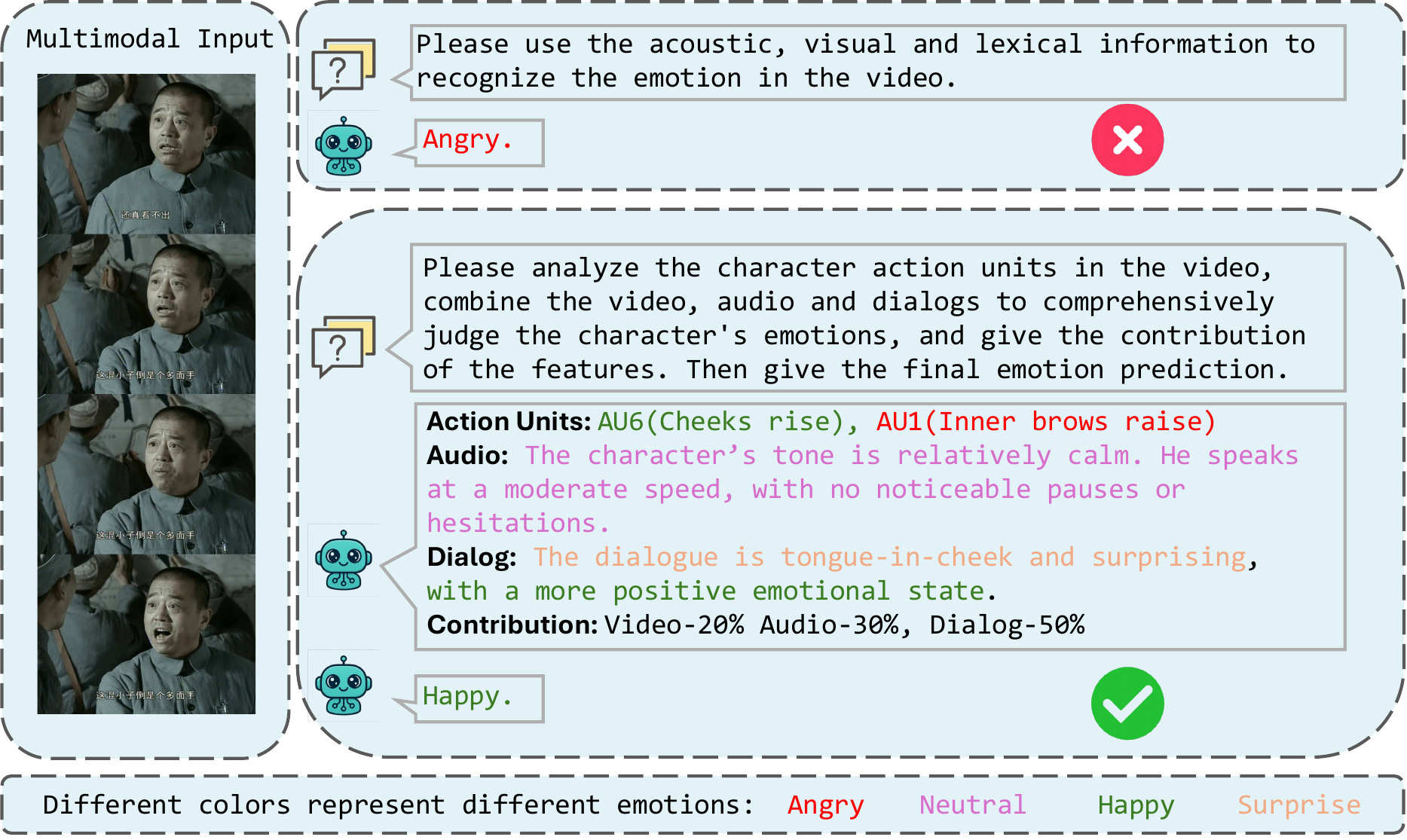}
    \caption{Reliable Reasoning Process}
    \label{fig:gemini}
\end{figure}

The whole reasoning priors can be expressed as
\begin{equation}
    P := \{c^* ~|~ I_V,I_A,I_T,R\}
\end{equation}
where $R$ denotes the contribution of each modality. For each modality prior, it may imply different emotion labels $c$ with a latent mapping $S(I_{m}):\rightarrow {c}$, MLLMs will synthesize all priors to recognize a certain emotion.
\begin{equation}
    c^* = \arg\max S(I_{m})
\end{equation}

In subsequent training stages, these trustworthy reasoning traces can be incorporated as priors guided dataset, guiding the multimodal fusion network and enhancing its integrative performance. 

\subsection{Trustworthy MultiModal Fusion Paradigm}
Annotating multimodal video–based emotion‐recognition corpora is intrinsically difficult, as automated labeling pipelines typically yield sub-par accuracy, whereas human verification is prohibitively expensive. As a result, these datasets characteristically contain a substantial proportion of unlabeled samples.

Each video segment can be decomposed into three complementary information streams—visual frame sequences, acoustic waveforms, and transcript-level dialogue text. After passing each stream through its dedicated modality-specific encoder, we obtain three heterogeneous feature vectors, which are subsequently projected and harmonised within a common latent space by a unified feature projector.

\begin{figure}
    \centering
    \includegraphics[width=0.9\linewidth]{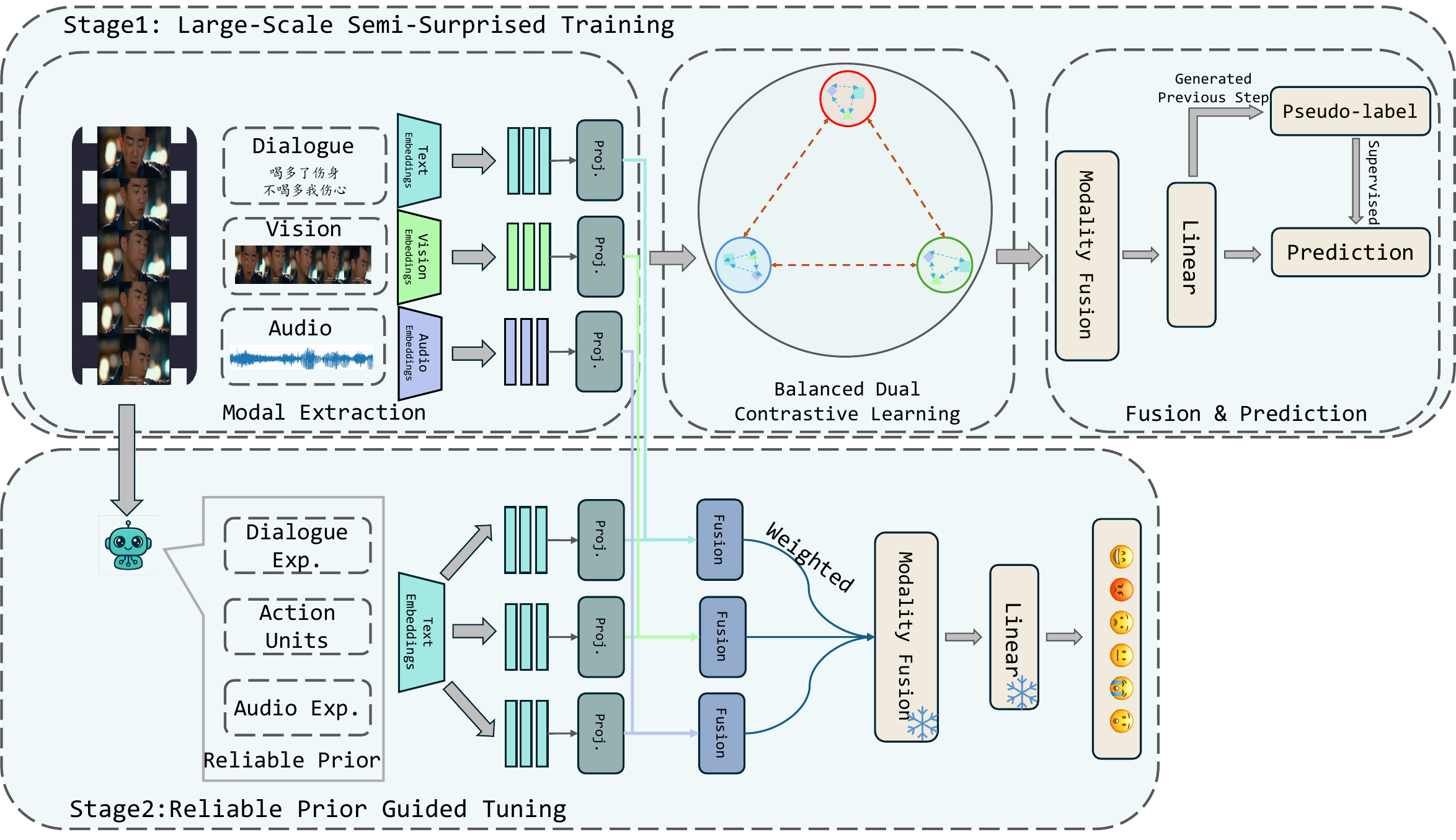}
    \caption{The whole framework of the multimodal fusion paradigm.}
    \label{fig:wkflow}
\end{figure}

Our multimodal fusion framework is optimised through a two-stage training regimen as Fig \ref{fig:wkflow}:

The first stage is \textbf{Large-scale Semi-supervised Pre-training}, which leverages both labeled and unlabeled data to learn robust cross-modal representation. And the second stage is \textbf{Reliable Prior Guided Tuning}, wherein domain-specific priors are incorporated to refine the model for downstream emotion-recognition tasks with enhanced reliability.

During the first stage of large-scale pre-training, our objective is to enforce intra-class compactness while simultaneously maximizing inter-class separability across all emotion categories. Therefore, we need to pull the feature vectors of the same category as close as possible, and pull the feature vectors of different categories as far as possible. The contrastive learning strategy of infoNCE can effectively achieve this separability. However, since the labels of the original expression samples are not balanced, we introduce

To that end, we employ a \textbf{Balanced Dual Contrastive Learning} (BDCL) strategy. In BDCL, we construct two parallel contrastive objectives: the First is inter-modality contrast, which draws together representations of the same video segment across different modalities, and the second is intra-modality contrast, which pushes apart representations originating from different emotion classes within the same modality. Formally, let

\begin{equation}
\mathbf{z}_{v},\;\mathbf{z}_{a},\;\mathbf{z}_{t}\in \mathbb{R}^{d},
\end{equation}
denote the projected features for the visual, audio, and textual streams of a given sample. For a mini-batch $\mathcal{B}$ containing $N$ labeled examples, we define the dual contrastive loss as
\begin{equation}
    \mathcal{L}_{\mathrm{BDCL}}
=\lambda_{\text{inter}}\;\mathcal{L}_{\text{inter}}(\mathcal{B})
+\lambda_{\text{intra}}\;\mathcal{L}_{\text{intra}}(\mathcal{B}),
\end{equation}
where the weighting hyper-parameters $\lambda_{\text{inter}}$ and $\lambda_{\text{intra}}$ balance the two components. 

The intra-modality term applies an emotion-aware InfoNCE~\cite{oord2018cpc} loss independently within each modality, using labels to regard features with the same emotion as positives and all others as negatives. For sample $i$ with modality $m$ with same-label positive index set $\mathcal{P}^{intra}_{m,i} = \{j \neq i | y_i=y_j\}$, the $\mathcal{L}_{\text{intra}}(\mathcal{B})$ is 

\begin{equation}
     \mathcal L_{\text{intra}}=\frac{1}{|\mathcal M|N} \sum_{m\in\mathcal M}\!\sum_{i=1}^{N}\! \frac{1}{|\mathcal P^{\text{intra}}_{m,i}|} \sum_{j\in\mathcal P^{\text{intra}}_{m,i}} -\log \frac{\exp\bigl((\mathbf z_{m,i}^\top~\mathbf z_{m,j})/\tau\bigr)} {\displaystyle \frac{1}{C}\sum_{c=1}^{C}\! \frac{1}{|\mathcal{B}^{(m)}_c|} \sum_{k\in \mathcal{B}^{(m)}_c} \exp\bigl((\mathbf z_{m,i}^\top~\mathbf z_{m,k})/\tau\bigr)} 
\end{equation}
where $|\mathcal M|$ represents the three modalities and $\mathcal{B}_c$ the index set of all modality embeddings in the batch whose label is $c$, $\mathcal{B}^{(m)}_c \subset \mathcal{B}_c $ restricts to embeddings of modality $m$. 

The inter-modality term is a modified symmetric InfoNCE loss that treats each pair as  $(m,n) \in \mathcal{P}^{inter}_i = \{(\mathbf{z}_{v},\mathbf{z}_{a}),(\mathbf{z}_{v},\mathbf{z}_{t}),(\mathbf{z}_{a},\mathbf{z}_{t})\}$ of the same sample as positives, while all remaining modality pairs in the batch serve as negatives. 

\begin{equation}
\mathcal L_{\text{inter}} =\frac{1}{|\mathcal P^{inter}|N} \sum_{i=1}^{N}\! \sum_{(m,n)\in\mathcal P^{\text{inter}}_i} -\log \frac{\exp\bigl((\mathbf z_{m,i}^\top\mathbf z_{n,i})/\tau\bigr)} {\displaystyle \frac{1}{C}\sum_{c=1}^{C}\! \frac{1}{|\mathcal{B}_c|} \sum_{k\in \mathcal{B}_c} \exp\bigl((\mathbf z_{m,i}^\top~\mathbf z_{k})/\tau\bigr)} 
\end{equation}

The class-balanced denominator
\begin{equation}
    D_{m,i}=\frac{1}{C}\sum_{c=1}^{C}\frac{1}{|\mathcal{B}_{c}|}\sum_{k\in\mathcal{B}_{c}}\exp\!\Bigl(\tfrac{\mathbf z_{m,i}^{\top}\mathbf z_{k}}{\tau}\Bigr)
\end{equation}
works because it implicitly replaces the usual “pick any negative from the batch”~\cite{zhu2022bcl} rule with a two-step procedure that first selects a class uniformly and then samples a representation uniformly from that class.  This yields a balanced negative distribution in which every category, no matter how rare in the real data, contributes exactly $1/C$ of the probability mass, as head classes no longer exert a disproportionately large repulsive force, and tail-class embeddings are prevented from collapsing toward the origin. 
\begin{equation}
\label{eq:grad}
\mathbb{E}\!\bigl[-\nabla_{\mathbf z_{m,i}}\,
\log D_{m,i}\bigr]
\;\propto\;
\frac{1}{C}\sum_{c=1}^{C}
\mathbb{E}_{k\sim c}\!
\bigl[\,
\nabla_{\mathbf z_{m,i}}\bigl(\mathbf z_{m,i}^\top
~\mathbf z_{k}\bigr)
\bigr], 
\end{equation}

Given the scarcity of annotated examples, the training corpus comprises a large proportion of unlabeled instances. After an initial cold-start phase trained solely on the labeled subset, the model iteratively assigns pseudo-labels to the unlabeled data at the conclusion of each epoch. These pseudo-labels are then incorporated as additional supervisory signals, enabling a semi-supervised optimization of the network. Upon completion of stage 1, the resulting model demonstrates a robust capacity for emotion recognition.

We use reliable emotion priors data generated by Gemini in \ref{sec3.1} as our stage 2 tuning dataset to distill high-level reasoning priors from MLLMs into a lightweight multi-modal recognition architecture.

In this stage, we refine reasoning priors into text, audio, and video modalities as
\begin{equation}
\begin{aligned}
        \mathbf{e_a} = \text{Proj}(\text{Embeddings}(I_A))\\
        \mathbf{e_v} = \text{Proj}(\text{Embeddings}(I_V))\\
        \mathbf{e_t} = \text{Proj}(\text{Embeddings}(I_T)).\\
\end{aligned}
\end{equation}

Then we can fuse them with corresponding modal information 
\begin{equation}
    \mathbf{F_i} = \text{Transformer}(\text{MLP}(\mathbf{e_i}+\mathbf{z_i})).
\end{equation}

In the process, we freeze the modality fusion and linear layer as well as the encoder of each modality, and only train the encoder and fusion used by the pre-introduction prior. They are composed of linear and transformer modules, and at the same time, the modality contribution ratio provided by the prior information is used as the modality fusion weight.


\section{Experiments}

\subsection{Experiments Setup}

The main experiments were conducted on the MER2024 dataset. We use MER2024~\cite{lian2024mer} labeled samples and partially MER2023~\cite{lian2023mer} test dataset provided by Emotion-Llama~\cite{cheng2024emotion} as our supervised training dataset. All training was done on a single NVIDIA A100. we use \textit{clip-vit-large-patch14, chinese-hubert-large, bloom-7b} as the feature embedding for visual, audio, and text, respectively.
We set $\lambda_{inter}$ and $\lambda_{intra}$ both to  0.2 in BDCL, and we follow the infoNCE conventional setting $\tau$ to 0.1.

\subsection{Main Result}

A comprehensive empirical comparison between the LLM-based method and the modality-fusion method is conducted on the MER2024-SEMI benchmark’s test partition, with the quantitative outcomes summarized in Table \ref{tab1}.
\begin{table}
  \centering
\caption{Comparison of different methods on MER2024 dataset. The corresponding indicators are the classification accuracy under each category label and the overall classification accuracy.}\label{tab1}
\begin{tabular}{|l|l|l|l|l|l|l|l|}
\hline
Method & Avg & Neutral & Angry & Happy & Sad & Worried & Surprise \\
\hline
LLM-based Method&&&&&&&\\
Qwen-Audio~\cite{chu2023qwenaudio}          & 31.74 & 67.04 & 29.20 & 25.97 & 12.93 & 35.36 &  6.12 \\
LLaVA-NEXT~\cite{li2024llavanext}      & 33.75 & 38.95 &  0.00 & 57.46 & 79.42 &  0.00 &  0.00  \\
MiniGPT-v2~\cite{chen2023minigptv2}   & 34.47 & 22.28 & 20.69 & 84.25 & 47.23 &  0.55 &  2.04  \\
Video-LLaVA(image)~\cite{lin2023videollava} & 31.10 & 26.97 & 58.85 & 37.09 & 27.18 &  3.31 & 12.97 \\
Video-LLaVA(video)~\cite{lin2023videollava} & 35.24 & 29.78 & 58.85 & 51.94 & 39.84 &  2.76 &  0.00  \\
Video-Llama~\cite{zhang2023videollama}   & 35.75 & 80.15 &  5.29 & 20.25 & 67.55 &  9.39 &  4.76  \\
GPT-4o                & 58.43 & 59.54 & 56.44 & 67.33 & 75.03 & 57.00 & 35.23 \\
Gemini-2.0-flash &71.81&53.97&82.46&91.70&92.18&49.54&61.02 \\
\hline
Modality-fusion Method&&&&&&&\\
Attention  & 77.53 & 75.10 & 76.11 & 91.82 & 85.31 & 64.57 & 32.43 \\
Contrastive & 79.72 & 76.68 & \textbf{82.74} & 87.73 & 86.01 & 70.08 & 43.24 \\
Ours&\textbf{84.68}&	\textbf{88.72}&	78.65&	\textbf{91.89}&	\textbf{88.44}&	\textbf{76.22}&	\textbf{70.83}\\
\hline
\end{tabular}
\end{table}

Although LLMs benefit from extensive pre-training on broad, general-domain corpora, they still exhibit marked domain shift when confronted with specialised tasks such as fine-grained emotion recognition. Closed-source systems mismatch to a degree and generally outperform open-source counterparts; however, a non-trivial performance gap remains relative to traditional, task-specific multimodal fusion architectures.

"Attention" reports the baseline scores provided by the MER2024 benchmark. And "Contrastive" presents the performance obtained after augmenting the architecture with a contrastive-learning objective. When robust domain priors are subsequently infused into this contrastive framework, the model exhibits a pronounced uplift across all evaluation metrics, indicating a substantial enhancement in overall performance.
\subsection{Ablations}

\begin{figure}
    \centering
    \begin{subfigure}[b]{0.48\linewidth}
        \includegraphics[width=\linewidth]{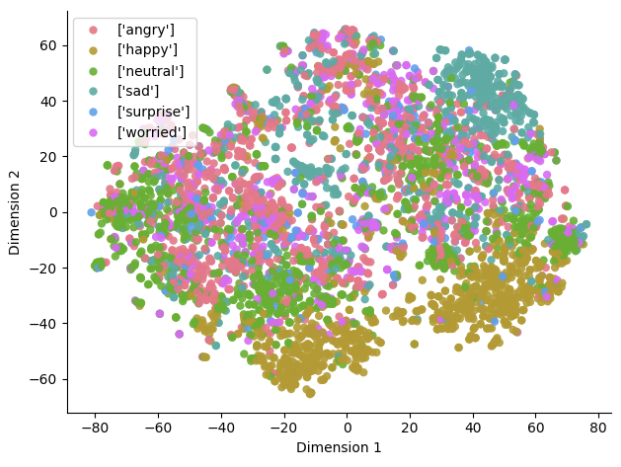}
        \caption{With BDCL}
        \label{fig:au-a}
    \end{subfigure}
    \hfill
    \begin{subfigure}[b]{0.48\linewidth}
        \includegraphics[width=\linewidth]{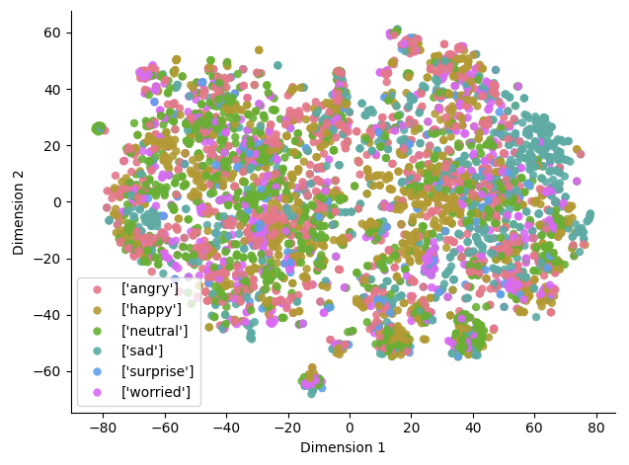}
        \caption{Without BDCL}
        \label{fig:au-b}
    \end{subfigure}
    \caption{Latent feature space comparison.}
    \label{fig:au}
\end{figure}

\paragraph{\textbf{Modality Ablations}}Table \ref{results} presents an ablation analysis that quantifies the individual contributions of each modality to the emotion-recognition task. When the model is restricted to a single input channel, the acoustic pathway consistently yields the highest performance. We attribute this superiority to the pronounced discriminative boundaries embedded in speech-spectral cues, which map affective states onto well-separated regions of the feature space. Moreover, the incorporation of our enhanced, reliability-weighted priors further amplifies this advantage, delivering additional gains in recognition accuracy.
\begin{table}
  \centering
\caption{\label{results}Results of different modalities.} 
\begin{tabular}{|l|l|l|}
\hline
Modality    & Accuracy    & F1-Score  \\
\hline
    AVT+ Prior     &84.68   & 84.70         \\
    AVT  & 77.53   & 77.25     \\
    A  & 74.16   & 74.10          \\
    V  & 66.30   & 66.64           \\
    T  & 51.89   & 51.93          \\
\hline
\end{tabular}
\end{table}
\paragraph{\textbf{Balanced Sample}} 
Table \ref{mer23} reports the performance obtained after augmenting the label set with MER2023 labeled samples from Emotion-LLama. Here \textit{random} indicates random sampling, \textit{matched} indicates sampling expansion according to the original label ratio, and \textit{balanced} indicates using samples to balance the original label ratio. The comparative results highlight how different balancing schemes modulate downstream learning dynamics and ultimately shape overall recognition accuracy.
\begin{table}
  \centering
\caption{\label{mer23}Results of different data samples.} 

\begin{tabular}{|l|l|l|l|}
\hline
Dataset & Size   & Accuracy & F1‑score  \\
\hline
    MER2024           & 4024   & 78.03   & 78.25         \\
    MER2024 +1K, random       & 5024     & 78.33   & 78.27         \\
    MER2024 +1K, matched    & 5022   & 78.73   & 78.69           \\
    MER2024 +1K, balanced   & 5024     & 77.34   & 77.38          \\
    MER2024 +4K, all       & 8341     & 79.22   & 79.19         \\
\hline
\end{tabular}
\end{table}
\paragraph{\textbf{Contrastive Learning Ablations}}
Figure \ref{fig:au} depicts the sample distributions in the latent feature space following t-SNE projection, comparing models trained with and without BDCL. In the contrastive-learning condition, the embeddings exhibit markedly tighter intra-class clusters and clearer inter-class separation, indicating that the training procedure promotes a more compact and discriminative feature representation.

\section{Conclusion}

In this study, we explored the integration of reliable reasoning priors from MLLMs into lightweight multimodal video emotion recognition frameworks.

Reliable priors not only enhance multimodal integration but also serve as a powerful signal for modality-specific feature refinement. To further address the challenge of label imbalance, we incorporate a balanced dual-contrastive learning strategy, which significantly improves sample separability and robustness. Empirical results validate the effectiveness of our approach across various emotion recognition scenarios.

%
%
%
%

{
\small
\bibliographystyle{unsrt}
\bibliography{ref}
}




\end{document}